\documentclass[jcp, twocolumn, floatfix, superscriptaddress, reprint, 10pt]{revtex4-1}
\usepackage[english]{babel}
\usepackage[utf8]{inputenc}
\usepackage[paperwidth=210mm,paperheight=297mm,centering,hmargin=2cm,vmargin=2.5cm]{geometry}
\usepackage[utf8]{inputenc}
\usepackage{verbatim}
\usepackage{bm}
\usepackage{amsmath}
\usepackage{amssymb}
\usepackage{natbib}
\usepackage{pdfpages}
\usepackage{graphics}
\usepackage{float}
\usepackage{hyperref}
\usepackage[capitalize]{cleveref}
\usepackage{multirow}
\usepackage{makecell}
\usepackage{outlines}
\usepackage{siunitx}
\usepackage{mhchem}
\usepackage{caption}
\usepackage{subcaption}
\usepackage{xcolor}
\usepackage{ulem}
\usepackage{comment}
\ifdefined\DIRACREP
\else
\newcommand{\DIRACREP}{}

\usepackage{physics}
\usepackage{xparse}
\ifdefined\COSMOMATHS
\else
\newcommand{\COSMOMATHS}{}

\newcommand{\mbf}[1]{\ensuremath{\mathbf{#1}}}

\newcommand{\D}[1]{\operatorname{d}{\!#1}\,}

\fi %

\NewDocumentCommand{\rep}{s d<| d|>}{%
\IfBooleanTF{#1}{
   \IfValueTF{#2}{
       \IfValueTF{#3}{\braket{#2}{#3}}{\bra{#2}}
       }{
       \IfValueTF{#3}{\ket{#3}}{}
       }
   }{
   \IfValueTF{#2}{
       \IfValueTF{#3}{\braket*{#2}{#3}}{\bra*{#2}}
       }{
       \IfValueTF{#3}{\ket*{#3}}{}
       }
   }
}

\NewDocumentCommand{\rbra}{sm}{\IfBooleanTF{#1}{\rep*<#2|}{\rep<#2|}}
\NewDocumentCommand{\rket}{sm}{\IfBooleanTF{#1}{\rep*|#2>}{\rep|#2>}}
\NewDocumentCommand{\rbraket}{smom}{
    \IfBooleanTF{#1}{
        \IfNoValueTF{#3}{\rep*<#2||#4>}{\rep*<#2|#3\rep*|#4>}
    }{
        \IfNoValueTF{#3}{\rep<#2||#4>}{\rep<#2|#3\rep|#4>}
    }
}

\NewDocumentCommand{\cg}{m m m}{\rep<#1; #2||#3>}

\NewDocumentCommand{\field}{o m e{_} e{^} o e{_} e{^}}{
\IfValueTF{#5}{\overline{
  #2\IfValueT{#3}{_#3}\IfValueT{#4}{^{\otimes #4}} %
  \otimes
  #5\IfValueT{#6}{_#6}\IfValueT{#7}{^{\otimes #7}} %
  \IfValueT{#1}{;#1}
}}{
  \IfValueTF{#4}{\overline{
     #2\IfValueT{#3}{_#3}\IfValueT{#4}{^{\otimes #4}}
     \IfValueT{#1}{;#1}
  }}
  {#2\IfValueT{#3}{_#3}}
}
}

\NewDocumentCommand{\frho}{o e{_} e{^}}{
\field[#1]{\rho}_{#2}^{#3}
}

\newcommand{\br}{\mbf{r}}
\newcommand{\bx}{\mbf{x}}
\newcommand{\bxhat}{\hat{\mbf{x}}}
\newcommand{\brhat}{\hat{\mbf{r}}}

\newcommand{\e}{a}  %

\NewDocumentCommand{\ex}{e_}{
\IfValueTF{#1}{\e_{#1}\bx_{#1}}{\e\bx}
}  %

\NewDocumentCommand{\lm}{e_}{
\IfValueTF{#1}{l_{#1}m_{#1}}{lm}
}
\NewDocumentCommand{\nlm}{e_}{
\IfValueTF{#1}{n_{#1}\lm_{#1}}{n\lm}
}
\NewDocumentCommand{\enlm}{e_}{
\IfValueTF{#1}{\e_{#1}\nlm_{#1}}{\e\nlm}
}
\NewDocumentCommand{\en}{e_}{
\IfValueTF{#1}{\e_{#1}n_{#1}}{\e n}
}
\NewDocumentCommand{\nlk}{e_}{
\IfValueTF{#1}{n_{#1}l_{#1}k_{#1}}{nlk}
}
\NewDocumentCommand{\enlk}{e_}{
\IfValueTF{#1}{\e_{#1}\nlk_{#1}}{\e\nlk}
}
\NewDocumentCommand{\enl}{e_}{
\IfValueTF{#1}{\en_{#1}l_#1}{\en l}
}

\NewDocumentCommand{\nnl}{s}{
\IfBooleanTF{#1}{n_1 n_2 l}{n_1; n_2; l}
}
\NewDocumentCommand{\ennl}{s}{
\IfBooleanTF{#1}{\en_1 \en_2 l}{\en_1; \en_2; l}
}

\NewDocumentCommand{\gslm}{s}{
\IfBooleanTF{#1}{\sigma\lambda\mu}{\sigma;\lambda\mu}
}

\newcommand{\Rhat}{\hat{R}}

\newcommand{\nmax}{n_\text{max}}
\newcommand{\lmax}{l_\text{max}}

\newcommand{\rcut}[0]{{r_\text{cut}} }

\fi %

\ifdefined\COSMOMODELS
\else
\newcommand{\COSMOMODELS}{}
\usepackage{bm,upgreek}

\newcommand{\feat}{\upxi}
\newcommand{\bfeat}[0]{\ensuremath{\bm{\upxi}}}

\fi %

\NewDocumentCommand\te{s}{\tilde{\e}\IfBooleanTF{#1}{'}{}}
\NewDocumentCommand\tn{s}{\tilde{n}\IfBooleanTF{#1}{'}{}}
\NewDocumentCommand\tl{s}{\tilde{l}\IfBooleanTF{#1}{'}{}}
\NewDocumentCommand\tm{s}{\tilde{m}\IfBooleanTF{#1}{'}{}}
\NewDocumentCommand\tlm{s}{\IfBooleanTF{#1}{\tl*\tm*}{\tl\tm}}
\NewDocumentCommand\tnlm{s}{\IfBooleanTF{#1}{\tnl*\tm*}{\tnl\tm}}
\NewDocumentCommand\tnl{s}{\IfBooleanTF{#1}{\tn*\tl*}{\tn\tl}}

\newcommand{\todorev}[1]{{}}
\makeatletter
\AtBeginDocument{\let\LS@rot\@undefined}
\makeatother
\begin{document}
\newcommand{\SM}{SM}
\newcommand{\rev}[1]{#1}
\makeatletter
\g@addto@macro\normalsize{%
  \setlength\abovedisplayskip{5pt}
  \setlength\belowdisplayskip{5pt}
  \setlength\abovedisplayshortskip{5pt}
  \setlength\belowdisplayshortskip{5pt}
  \setlength{\belowcaptionskip}{-10pt}
}
\makeatother

\setcitestyle{super}

\title{Unified theory of atom-centered representations \\ and \rev{message-passing} machine-learning schemes}
\author{Jigyasa Nigam}
\affiliation{Laboratory of Computational Science and Modeling, Institute of Materials, \'Ecole Polytechnique F\'ed\'erale de Lausanne, 1015 Lausanne, Switzerland}
\affiliation{National Centre for Computational Design and Discovery of Novel Materials (MARVEL), {\'E}cole Polytechnique F{\'e}d{\'e}rale de Lausanne, 1015 Lausanne, Switzerland}

\author{Sergey Pozdnyakov}
\affiliation{Laboratory of Computational Science and Modeling, Institute of Materials, \'Ecole Polytechnique F\'ed\'erale de Lausanne, 1015 Lausanne, Switzerland}

\author{Guillaume Fraux}
\affiliation{Laboratory of Computational Science and Modeling, Institute of Materials, \'Ecole Polytechnique F\'ed\'erale de Lausanne, 1015 Lausanne, Switzerland}

\author{Michele Ceriotti}
\email{michele.ceriotti@epfl.ch}
\affiliation{Laboratory of Computational Science and Modeling, Institute of Materials, \'Ecole Polytechnique F\'ed\'erale de Lausanne, 1015 Lausanne, Switzerland}
\affiliation{National Centre for Computational Design and Discovery of Novel Materials (MARVEL), {\'E}cole Polytechnique F{\'e}d{\'e}rale de Lausanne, 1015 Lausanne, Switzerland}

\onecolumngrid
\newcommand{\mc}[1]{{\color{purple}{#1}}}
\newcommand{\jn}[1]{{\color{red}{#1}}}
\newcommand{\gf}[1]{{\color{orange}{#1}}}

\begin{abstract}

Data-driven schemes that associate molecular and crystal structures with their microscopic properties share the need for a concise, effective description of the arrangement of their atomic constituents. 
Many types of models rely on descriptions of atom-centered environments, that are associated with an atomic property or with an atomic contribution to an extensive macroscopic quantity. Frameworks in this class can be understood in terms of atom-centered density correlations (ACDC), that are used as a basis for a body-ordered, symmetry-adapted expansion of the targets.
Several other schemes, that gather information on the relationship between neighboring atoms using ``message-passing'' ideas, cannot be directly mapped to correlations centered around a single atom. 
We generalize the ACDC framework to include multi-centered information, generating representations that provide a complete linear basis to regress symmetric functions of atomic coordinates, and provides a coherent foundation to systematize our understanding of both atom-centered and message-passing, invariant and equivariant machine-learning schemes.

\end{abstract}

\twocolumngrid

\maketitle

\begin{figure}[b]
    \centering
    \includegraphics[width=1.0\linewidth]{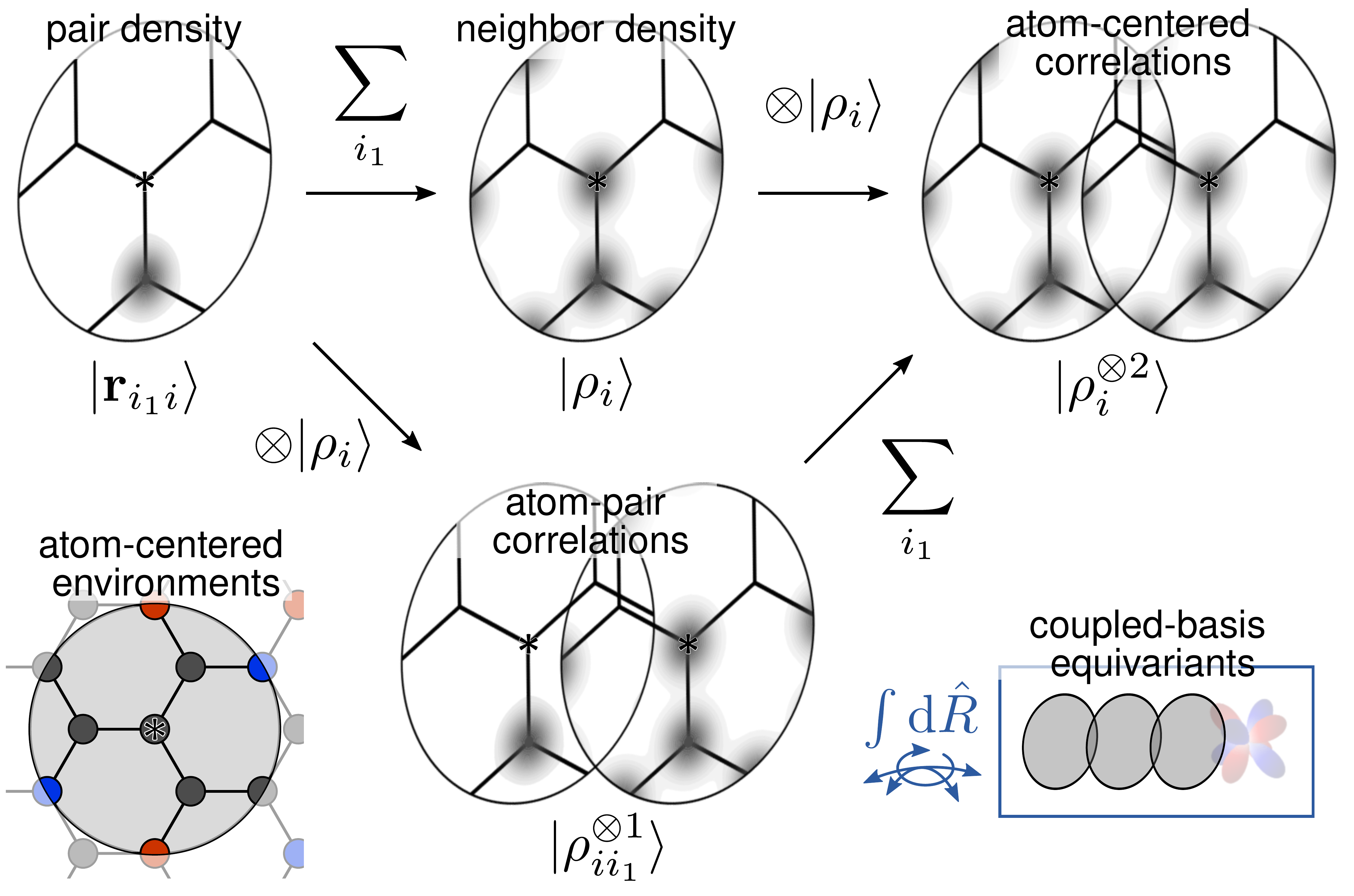}
    \caption{A few examples of ACDC representations, and their construction by summing pair contributions over atomic neighbors.}
    \label{fig:tensor-scheme}
\end{figure}

The understanding that $\nu$-point correlations of the atom density provide a complete linear basis to expand any property of these atom-centered environments\cite{shap16mms,will+19jcp,drau19prb} sheds light on the relations between different machine-learning (ML) frameworks for atomistic modeling,\cite{musi+21cr,uhri21prb,langer2020representations} and is driving much progress in obtaining more accurate and performant schemes\cite{zuo+20jpcl}.
This systematic approach is somewhat disconnected from efforts that combine ideas from geometric ML\cite{bronstein2021geometric, battaglia2018relational} and point clouds\cite{thomas2018tensor, fuchs2020se}, that look at molecules as graphs in which the atoms play the role of nodes, and the interatomic distances decorate the edges that connect them\cite{gilm+17icml, schutt2017dtnn, xie2018crystal, lubbers2018hierarchical, schutt2018schnet, unke2019physnet} -- a point of view that has a long-standing tradition in the field of cheminformatics\cite{david2020molecular, kear+16jcamd}.
\rev{Message-passing (MP) concepts, that describe the propagation of information about the geometric relations between nodes in terms of messages propagated along the edges, provide a general framework that encompasses popular architectures such as graph convolutional neural networks (NN) as special cases. }
With the rise of equivariant neural networks\cite{kondor2018covariant,thomas2018tensor, kondor2018clebsch,anderson2019cormorant,batzner2021se3,qiao2021unite} the similarity between atom-centered and message-passing schemes has become more and more apparent, since both rely on symmetry-adapted constructions based on O(3)-preserving operations. 
We introduce a formalism to define representations which incorporate graph-based and message-passing ideas, provide a linear basis to expand symmetric functions of the relative position of atoms (or points in a 3D cloud) and which correspond, in the appropriate limit, to a wide variety of deep-learning schemes based on graphs.
We give examples of the performance of these ``message-passing'' representations in describing short- and long-range structure-property relations, and discuss how this framework provides a unified formalism to describe geometric machine learning approaches that rely on structured representations\cite{battaglia2018relational}, incorporating inductive biases in the form of atom-centered contributions that are local and equivariant.

\begin{table}
\rev{
\caption{Notation key for different types of density-based representations\label{tab:notation-key}}
\renewcommand\arraystretch{1.6}
\begin{tabular}{l l}
\hline\hline
\multicolumn{2}{l}{Density peaked at the interatomic spacing $\br_{i_k}-\br_i$}\\
$\rep|\br_{i_k i}>$   & $\rep<\bx||\br_{i_k i}> = g(\bx-\br_{i_k i})$ \\\hline
\multicolumn{2}{p{0.95\linewidth}}{Basis of spherical-harmonics and radial functions}\\
$\rep<nlm|$   & $\equiv \int \D{\bx} \rep<n||x> \rep<lm||\bxhat> \rep<x\bxhat|$ \\\hline
\multicolumn{2}{l}{Neighbor density around the $i$-th atom}\\
$\rep|\rho_i>$   & $\equiv\ \sum_{i_1} \rep|\br_{i_1 i}>$    \\\hline
\multicolumn{2}{l}{ $\nu$-neighbor ACDC around the $i$-th atom}\\
$\rep|\rho_i^{\otimes \nu}>$   & $ \equiv\ \rep|\rho_i> \otimes \rep|\rho_i> \cdots \otimes \rep|\rho_i> $ \\\hline
\multicolumn{2}{p{0.95\linewidth}}{ \emph{Symmetrized} $\nu$-neighbor ACDC invariant}\\
$\rep|\frho_i^{\nu}>$   & $\equiv\ \int \D{\Rhat} \rep|\rho_i^{\otimes \nu}> $ \\\hline
\multicolumn{2}{p{0.95\linewidth}}{Symmetrized $\nu$-neighbor ACDC equivariant with SO(3) character $\lambda$ and inversion character $\sigma$}\\
$\rep|\frho[\sigma; \lambda\mu]_i^{\nu}>$   & $\equiv\  \int \D{\Rhat} \Rhat\rep|\sigma> \otimes \Rhat\rep|\lambda\mu>\otimes \Rhat\rep|\rho_i^{\otimes \nu}> $ \\\hline
\multicolumn{2}{p{0.95\linewidth}}{Multi-center ACDC features describing a cluster ${i,i_1,\ldots i_N}$ 
of $N+1$ atoms }\\
$\rep|\rho_{ii_1\ldots i_N}^{\otimes \nu\nu_1\ldots\nu_N}>$   & $\equiv \rep|\rho_i^{\otimes \nu}> \otimes\rep|\rho_{i_1}^{\otimes \nu_1}> \otimes \rep|\br_{i_1 i}>$ \\\hline
\multicolumn{2}{p{0.95\linewidth}}{Simplest form of an $i$-centered message-passing ACDC representation}\\
$\rep|\rho_{i}^{\otimes [\nu \leftarrow \nu_1]}>$  & $\equiv\ \sum_{i_1} \rep|\rho_{ii_1}^{\otimes \nu \nu_1}>$  \\\hline
\multicolumn{2}{p{0.95\linewidth}}{Higher-order MP contraction }\\
$\rep|\rho_{i}^{\otimes [\nu \leftarrow (\nu_1, \nu_2)]}>$  & $\equiv\ \sum_{i_1 i_2} \rep|\rho_{ii_1i_2}^{\otimes \nu \nu_1 \nu_2}>$ \\\hline
\multicolumn{2}{l}{Iterating the MP construction}\\
$\rep|\rho_i^{\otimes [\nu \leftarrow [\nu_1 \leftarrow \nu_2]]}> $ & $\equiv\ \sum_{i_1} \rep|\rho_i^{\otimes \nu}> \otimes \rep|\rho_{i_1}^{\otimes [\nu_1 \leftarrow \nu_2]}>\otimes \rep|\br_{i_1 i}>  $ \\\hline\hline
\end{tabular}
}
\end{table}

\vspace{-2ex}
\section{Theory}

\vspace{-2ex}
\subsection{Atom-centered density correlations.}
The basic ingredient in the atom-centered density correlation (ACDC) framework\cite{will+19jcp} involves localized functions (e.g. Gaussians $g$ or Dirac $\delta$ distributions) centered at the atomic positions $\br_i$. Following the notation formalized in Section~3.1 of Ref.~\citenum{musi+21cr}, we indicate these atomic contributions as $\rep<\bx||\br_i> \equiv g(\bx -\br_i)$ (or $\delta(\bx -\br_i)$).  \rev{The notation is summarized in Table~\ref{tab:notation-key}, and some of the key quantities are illustrated schematically in Fig.~\ref{fig:tensor-scheme}.}
Repeated symmetrization, and summation over the neighbors of the central atom, yield a family of $(N+1)$-centers, $(\nu+1)$-body-order equivariant representations,\cite{niga+22jcp} 
\begin{equation}
\!\!\!\!\rep|\frho[\sigma; \lambda\mu ]_{{ii_1\ldots i_N}}^{\nu}> \equiv
\!\!\int_{O(3)}\!\!\!\!\!\!\!\!\! \D{\Rhat} \Rhat\! \rep|\lambda\mu> \!\otimes\! \Rhat\!\rep|\sigma> \!\otimes\! 
\Rhat\!\rep|\rho_{{ii_1\ldots i_N}}^{\otimes \nu}>.
\label{eq:ncenter-slm}
\end{equation} 
In this expression, $\rep|\lambda\mu>$ is a vector that tracks the \rev{irreducible spherical (SO(3)) representation \cite{morri-park87ajp}} and, $\rep|\sigma>$, the parity label. $\rep|\rho_{{ii_1\ldots i_N}}^{\otimes \nu}>$ indicates a tensor product of pair terms $\rep|\br_{i_k i}>$, that determine the relative position $\br_{i_ki}=\br_{i_k} - \br_i$ of the $N$ centers around the central atom, and of $\nu$ of the neighbors found within the $i$-centered environment $A_i$ \rev{(for a derivation, see Ref.~\citenum{niga+22jcp}, where the following is presented as Eq. (5))}
\begin{equation}
\!\!\rep|\rho_{{ii_1\ldots i_N}}^{\otimes \nu}> \!=\!
\rep|\br_{i_1 i }> \otimes \cdots \!\rep|\br_{i_N i}> \otimes \!\!\!\!\!\!\! \sum_{j_1\ldots j_\nu\in A_i} \!\!\!\!\!\!\!\! \rep|\br_{j_1i}> \otimes \cdots \! \rep|\br_{j_\nu i}>.
\label{eq:ncenter-ket}
\end{equation}
Eq.~\eqref{eq:ncenter-ket} is a highly abstract formulation of the construction (represented schematically in Fig.~\ref{fig:tensor-scheme}), that can be evaluated in practice by projection on any complete basis $\rep|q>$. 
By combining the resulting coefficients with appropriately fitted weights $\rep<y||q>$, one can approximate any $(N+1)$-center quantity $y_{ii_1\ldots i_N}$ that is symmetric with respect to permutations of the labels of the neighbors, and that is equivariant with respect to rigid translations and rotations, e.g. 
\begin{equation}
y_{ii_1\ldots i_N}^{\sigma;\lambda\mu}(A) \approx
\sum_q \rep<y; \sigma; \lambda||q> \rep<q||\frho[\sigma;\lambda\mu]_{{ii_1\ldots i_N}}^{\nu}>.
\end{equation}
To see how, one can consider a real-space basis and a $\delta$-like atom density. For an invariant (scalar) property, 
\begin{multline}
y_{ii_1\ldots i_N}(A) = 
\int\D\{\bx\} \rep<y||\{\bx\}>\rep<\{\bx\}||\field{\delta}_{{ii_1\ldots i_N}}^{\nu}>
=\\[-1ex]
\!\!\int_{O(3)}\!\! \!\!\!\!\D{\Rhat}  \!\!\!\!\sum_{j_1\ldots j_\nu\in A_i}  \!\!\!\!
\tilde{y}(\Rhat\br_{j_1i}, \ldots \Rhat\br_{j_\nu i}, \Rhat\br_{i_1 i}, \ldots \Rhat\br_{i_N i}),
\label{eq:ncenter-model}
\end{multline}
where $\{\bx\}$ is a shorthand for the collection of the continuous indices associated with each interatomic degree of freedom. \footnote{A derivation for the $N=1$ case, that follows closely the ideas in Ref.~\citenum{shap16mms}, can be found in Section 4.1 of Ref.~\citenum{musi+21cr}.}

\begin{figure}[btp]
    \centering
    \includegraphics[width=1.0\linewidth]{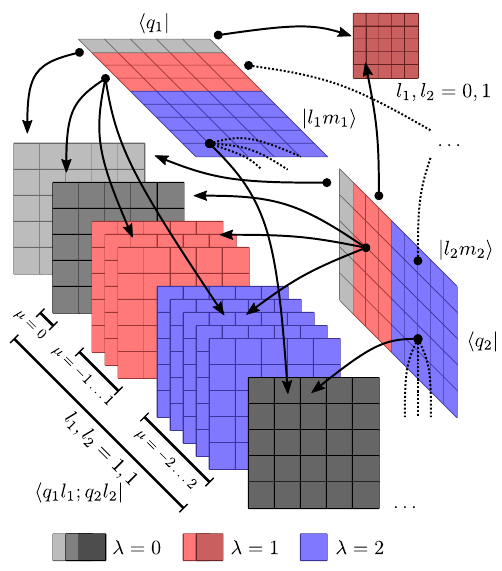}
\caption{\rev{Equivariant features (color-coded based on the SO(3) character, and enumerated by the index $q$) can be combined to generate equivariants of higher order by CG iterations (black arrows). The figure only shows a small part of the equivariants that can be computed, but emphasizes that linearly independent terms with the same SO(3) character arise from combination of different characters, giving rise to a richer product space $\rep<q_1l_1; q_2l_2|$. \label{fig:cg-iter}
} }    
\end{figure}
Explicitly evaluating the sum over neighbors entails a cost that scales exponentially with $\nu$. However, a \emph{density trick} eliminates the steep scaling with the number of neighbors, by first computing an atom-centered density,  $\rep|\rho_i> \equiv \sum_j \rep|\br_{ji}>$, and then performing the tensor products.\cite{musi+21cr} 
Similarly, one does not need to compute the integral over rotations explicitly. By expressing the equivariant features in an irreducible \emph{coupled} basis, and discretizing each term in the tensor product on a basis of radial functions $\rep<x||nl>$ and spherical harmonics $\rep<\hat{\bx}||lm>$, the full set of symmetry-adapted equivariants~\eqref{eq:ncenter-slm} can be evaluated by combining symmetry-adapted terms in an iterative fashion\cite{niga+20jcp}
\begin{multline}
\rep<q_1 l_1 ; q_2 l_2 ||A_1; A_2; (\sigma_1\sigma_2(-1)^{l_1+l_2+\lambda}); \lambda\mu> = \\
\!\!\!\sum_{m_1 m_2}\!\!\! \rep<q_1|| A_1; \sigma_1; l_1 m_1> \! \rep<q_2||A_2; \sigma_2; l_2 m_2> \!\cg{l_1 m_1}{l_2 m_2}{\lambda \mu }.\\[-2ex]\label{eq:gen-cg-iter}
\end{multline}

\rev{The key insight from Eq.~\eqref{eq:gen-cg-iter} is that equivariant features can be combined to yield higher-order equivariants that provide a more complete basis to express structure-property relations. 
As shown graphically in Figure~\ref{fig:cg-iter}, the tensor product of two features with symmetry $l_1$ and $l_2$ generates equivariant components ranging from $\lambda=|l_1-l_2|$ to $l_1+l_2$.}
Expressions such as~\eqref{eq:ncenter-ket} provide a notation to describe a linear basis to approximate properties of $(N+1)$ atomic centers that depend simultaneously on the positions of $\nu$ neigbours, and~\eqref{eq:gen-cg-iter} a strategy to compute such features iteratively, while keeping the irreducible equivariant components separate. 
\rev{ Refs.~\citenum{glie+18prb,will+19jcp,drau19prb} discuss the universality of this construction, based on its equivalence to a complete basis of symmetric polynomials (see also Section 4.1 of Ref.~\citenum{musi+21cr} for a pedagogic discussion). Similar arguments about approximating equivariant functions with a basis of equivariant polynomials have been used in Ref.~\citenum{dym-maro20arxiv} to prove the universality of tensor field networks.\cite{thomas2018tensor} 
}

\rev{
As a concrete example, consider the expansion of the atom-centered density in radial functions $\rep<r||n>$ and spherical harmonics $\rep<\brhat||lm>$,
$\rep<nlm||\rho_i>$, which can also be interpreted as the $\nu=1$ equivariant ACDC $\rep<n||\frho[lm]_i^{1}>$. 
The two-point correlation can be simply indexed as $\rep<nlm; n'l'm'||\rho_i^{\otimes 2}>$, but its elements do not transform as irreducible representations of O(3). 
We can obtain a more transparent equivariant formulation by using the iteration~\eqref{eq:gen-cg-iter},
\begin{multline}
\rep<nl; n'l'||\frho[(-1)^{l+l'+\lambda};\lambda\mu]_i^2>=\\
\sum_{mm'}
\rep<n||\frho[lm]_i^{1}>
\rep<n'||\frho[l'm']_i^{1}>
\cg{lm}{l'm'}{\lambda\mu},
\end{multline}
that yields $\lambda$-SOAP features\cite{gris+18prl} with a well-defined parity and SO(3) character. 
The $(\sigma=0, \lambda=0)$ scalar term corresponds to the popular invariant SOAP representation\cite{bart+13prb}
\begin{equation}
\rep<n n'l||\frho_i^2>=
\sum_{m}
\rep<n||\frho[lm]_i^{1}>
\rep<n'||\frho[lm]_i^{1}>.
\end{equation}
To better understand the broader significance of this construction, and its relationship to other geometric ML schemes, is worth recalling that SOAP invariants encode information on the distances between the central atom and two neighbors $r_{ji}, r_{j'i}$, as well as on the angle between them $\theta_{jj'i}$, and could be equally well computed as\cite{will+19jcp,musi+21cr}
\begin{equation}
\rep<n n'l||\frho_i^2>=
\sum_{j,j' \in A_i} G_{nn'l}(r_{ji}, r_{j'i}, \theta_{jj'i}),
\end{equation}
where the $G_{nn'l}$ are functions that depend on the basis used for the neighbor density expansion, and that could in principle be chosen to mimic the usual Behler-Parrinello symmetry functions\cite{behl11jcp}. 
Similar expressions can be derived for any scalar contraction of two atom-centered density terms: for instance, the $3$-centers, $0$-neighbors representation can be evaluated as 
\begin{multline}
\rep<n n'l||\frho_{{ii_1i_2}}^0>=
\sum_m \rep<nlm||\br_{i_1i}>
\rep<n'lm||\br_{i_2i}> \\[-2ex] 
=G_{nn'l}(r_{i_1i}, r_{i_2i}, \theta_{i_1i_2 i}),\label{eq:nu0-n2}
\end{multline}
that we will use later to discuss the connection with geometric ML schemes based on angular information. 
}

\begin{figure}[btp]
    \centering
    \includegraphics[width=1.0  \linewidth]{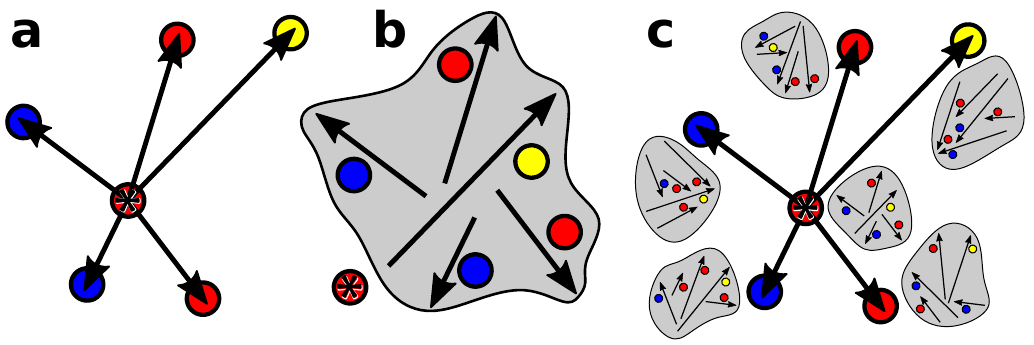}
    \caption{Schematic representation of the functioning of message-passing schemes. \rev{ The three panels illustrate the different steps of a message-passing architecture, as discussed in the text.} }
    \label{fig:gc-scheme}
\end{figure}

\subsection{\rev{Equivariant message-passing} representations.}
The ACDC construction is entirely based on terms centered around a single atom, and can be seen as alternative to message-passing (MP) architectures, which are based on the idea that information on multiple nodes (atoms, in this context) and their connectivity (inter-atomic distance vectors) could be combined to provide higher descriptive power.
\rev{The essential ideas behind MP frameworks can be summarized in terms of a few essential steps, illustrated in Fig.~\ref{fig:gc-scheme}: (a) A description of the connectivity of a selected node $i$ with each of its neighbors $j$ is built by combining information on the neighbor attributes, and on the geometry of the connection; (b) Information on all the neighbors of the selected node is combined in a way that is independent on the order by which they are considered; (c) The compound description of the neighborhood is assigned as the new attributes of the central node. Iterating the process propagates information on the neighbors' neighbors to every node. 
Graph-convolutional frameworks can be seen as a special case.\cite{bronstein2021geometric, battaglia2018relational}
} 
We refer the reader to Refs.~\citenum{gilm+17icml,schutt2019quantum} for an overview of the methodology applied to chemical modelling, \rev{and to section~\ref{sub:deep-l} for an explicit comparison with existing MP architectures.} 

The ACDC framework can be extended to provide a basis for the construction of arbitrarily complex equivariant MP frameworks.
First, we introduce the possibility of decorating an $(N+1)$-center representation with information on \emph{any} of the centers. 
Disregarding the symmetry indices and the integral over rotations for simplicity --- as discussed above, symmetry-adapted equivariants can be obtained by using Eq.~\eqref{eq:gen-cg-iter} on the tensor-product pattern --- we define the decorated atom-density representations
\begin{equation}
\rep|\rho_{{ii_1\ldots i_N}}^{\otimes \nu\nu_1 \ldots \nu_N}>\equiv \rep|\rho_i^{\otimes \nu}> \otimes  \rep|\rho_{i_1}^{\otimes \nu_1}>\otimes \rep|\br_{i_1 i}> \cdots  \otimes  \rep|\rho_{i_N}^{\otimes \nu_N}>\otimes \rep|\br_{i_N i} >.
\end{equation}
The $(N+1)$-center representations~\eqref{eq:ncenter-ket} are a special case, with $
\nu_{k>0}=0$. Given that we encode information on the \emph{vectors} connecting the central atom $i$ with its neighbors $\br_{i_k i}$, the information on the position of the neighbors relative to each other  is also formally retained, because $\br_{i_{k'}i_{k}}=\br_{i_{k'}i}-\br_{i_{k}i} $.
\rev{As we will show in section~\ref{sub:methane}, for a finite basis the performance of representations that formally describe the same order of interatomic correlations can depend, in practice, on which pairs of atoms are included explicitly in the description.}
\rev{
Taking tensor products of $(N+1)$-center features amounts to increasing their body order:
\begin{equation}
\!\!\!\!\rep|\rho_{{ii_1\ldots i_N}}^{\otimes \nu\nu_1 \ldots \nu_N}>\otimes \rep|\rho_{{ii_1\ldots i_N}}^{\otimes \nu'\nu'_1 \ldots \nu'_N}> \approx \rep|\rho_{{ii_1\ldots i_N}}^{\otimes (\nu+\nu')(\nu_1+\nu_1') \ldots (\nu_N+\nu_N')}>,
\end{equation}
where we use an approximate equality because on the left-hand side each $\rep|\br_{i_k i}>$ appears twice.  
The tensor product $\rep|\br_{i_k i}>\otimes \rep|\br_{i_k i}>$ increases the dimensionality of the left-hand side, even though it does not incorporate information on more neighbors or centers. 
}

\begin{figure}[btp]
    \centering
    \includegraphics[width=1.0\linewidth]{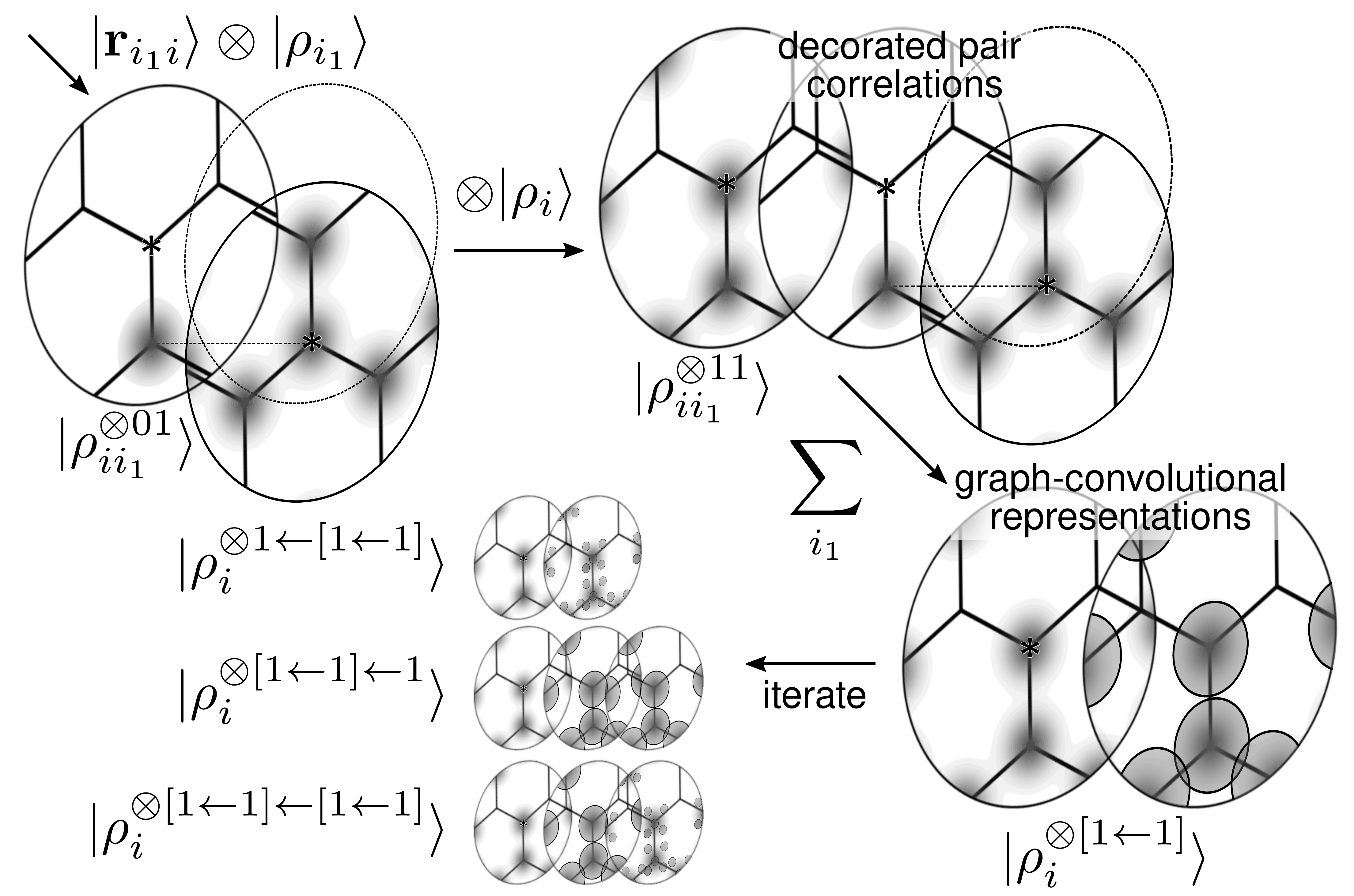}
    \caption{Construction of MP features from the contraction of pair and neighbor-density terms. The process can be iterated in many different ways (see the \SM{} for details).}
    \label{fig:tensor-gc}
\end{figure}

Simple MP-like representations can be obtained starting from decorated pair (2-center) features, and summing over the pair index (Fig.~\ref{fig:tensor-gc}):
\begin{equation}
\!\!\!\!\!\rep|\rho_{{i}}^{\otimes [\nu\leftarrow \nu_1]}>\!\equiv\! 
\sum_{i_1 \in A_i} \!\rep|\rho_{{ii_1}}^{\otimes \nu \nu_1}> \!=\! \rep|\rho_i^{\otimes \nu}> \!\otimes\!\! \sum_{i_1\in A_i} \! \rep|\rho_{i_1}^{\otimes \nu_1}> \!\otimes \!\rep|\br_{i_1 i}>.\label{eq:mp-simple}
\end{equation}
For the case of $\nu_1= 0$, the sum simplifies to $\rep| \rho_i>$, thus the expression simply evaluates ACDC features of order $\nu+1$. 
By expanding the tensor product, and in analogy with~\eqref{eq:ncenter-model}, one sees that these features provide a linear basis to expand a symmetrized function of the neighbors of $i$ and of each of their neighbors, e.g.
\begin{equation}
\sum_q \rep<y||q>\rep<q||\rho_{{i}}^{\otimes [1\leftarrow 1]}>=
\!\!\!\!\!\!\!\!\sum_{i_1,j\in A_i; j_1 \in A_{i_1}} \!\!\!\!\!\!\!\!
\tilde{y}(\br_{ji}, \br_{i_1i}, \br_{j_1 i_1}).
\label{eq:gc-11-model}
\end{equation}
A basis for O(3)-equivariant predictions can be obtained by computing the tensor product iteratively in the coupled basis, following~\eqref{eq:gen-cg-iter}.
Except for the (important) subtlety that one of the indices extends over the neighborhood of $i_1$, these features contain the same information as the $\nu=3$ ACDC (the bispectrum). Indeed, one can verify that $\rep|\frho_i^{[1\leftarrow1]}>$ discriminate between structures that are degenerate~\cite{pozd+20prl} for the on-site powerspectrum $\rep|\frho_i^2>$, but cannot discriminate between environments that have the same $\rep|\frho_i^3>$. 

Eq.~\ref{eq:mp-simple} can be generalized in different ways. For example, one could compute $N=2$ decorated features and contract them
\begin{equation}
\rep|\rho_{i i_1}^{\otimes [\nu\leftarrow \nu_2]\nu_1}>\equiv 
\sum_{i_2} \rep|\rho_{{ii_1i_2}}^{\otimes \nu \nu_1 \nu_2}>.\label{eq:mp-n3}
\end{equation}
These features describe atomic pairs as a sum over descriptors of triplets. They could be used as features to characterize edges, and seen as a building block of MP architectures that use bonds as nodes and angles as edges\cite{choudhary2021atomistic}.

By also summing over $i_1$, one obtains message-passing ACDC features that contain joint information on two decorated neighbors:
\begin{equation}
\label{eq:contract-two-centers}
\!\rep|\rho_i^{\otimes [\nu\leftarrow(\nu_1,\nu_2)]}> \!=\! \sum_{i_1 i_2}\!
\rep|\rho_i^{\otimes\nu}>\otimes\! \rep|\br_{i_1 i}>\otimes\! \rep|\rho_{i_1}^{\otimes\nu_1}>
\otimes\! \rep|\br_{i_2 i}> \otimes\! \rep|\rho_{i_2}^{\otimes\nu_2}>.
\end{equation}
This corresponds to an overall correlation order of $\nu+\nu_1+\nu_2+2$ neighbors around the the $i$-atom. More generally, representations that incorporate joint information on multiple neighbors can be built starting from pair MP features --- see the supplementary material (\SM) for more information:
\begin{equation}
\rep|\rho_{{i}}^{\otimes [\nu\leftarrow \nu_1]}>\otimes \rep|\rho_{{i}}^{\otimes [\nu'\leftarrow \nu_1']}> =
\rep|\rho_i^{\otimes [(\nu+\nu')\leftarrow(\nu_1,\nu_1')]}>.
\end{equation}
Note that the left-hand side avoids the double sum over neighbors in Eq.~\eqref{eq:contract-two-centers}, similar to how the density trick avoids summation over tuples of neighbors by first expanding a neigbor density, and then increasing the body order by taking tensor products rather than by explicit higher-order summations. 

Given that all of these contracted expressions yield representations that are \emph{atom-centered}, the process can be iterated. For example, one could first compute a low-order $\rep|\rho_i^{\otimes\nu}>$ descriptor, use it to evaluate the MP representation $\rep|\rho_i^{\otimes[\nu\leftarrow \nu]}>$, and then repeat. Depending on how one describes the central atom and the neighbors, the procedure yields different  types of representations.
\rev{
Combining atom-centered features on the $i$ atom with MP representations on the neighbors yields
\begin{equation}
\rep|\rho_i^{\otimes [\nu \leftarrow [\nu\leftarrow \nu]]}>=\sum_{i_1} \rep|\rho_i^{\otimes \nu}> \otimes \rep|\br_{i_1 i}> \otimes \rep|\rho_{i_1}^{\otimes [\nu \leftarrow \nu] }>, \label{eq:mp2-type-a}
\end{equation}
while combining MP representations on $i$ and atom-centered neighbor features yields
\begin{equation}
\rep|\rho_i^{\otimes [[\nu\leftarrow \nu]\leftarrow \nu]}>=\sum_{i_1} \rep|\rho_{i}^{[\nu\leftarrow \nu]}> \otimes \rep|\br_{i_1 i}> \otimes \rep|\rho_{i_1}^{\otimes \nu}>,
\label{eq:mp2-type-b}
\end{equation}
which can be shown to be equivalent to $\rep|\rho_i^{\otimes [\nu\leftarrow(\nu,\nu)]}>$.
Finally, one can combine MP representations on both atoms
\begin{equation}
\!\!\!\rep|\rho_i^{\otimes [[\nu\leftarrow \nu]\leftarrow [\nu\leftarrow \nu ]]}>=\sum_{i_1} \rep|\rho_{i}^{\otimes[\nu\leftarrow \nu]}> \otimes \rep|\br_{i_1 i}> \otimes \rep|\rho_{i_1}^{\otimes[\nu\leftarrow \nu]}>,
\label{eq:mp2-type-c}
\end{equation}
which can also been rewritten in terms of~\eqref{eq:mp2-type-a}.
A full expansion of these expressions is given in the \SM{}.}
Each of these descriptors -- as well as countless variations in which primitive features of different body order, describing atoms, bonds or general clusters, are combined together -- is associated with a tensor product string in the form of Eq.~\eqref{eq:ncenter-ket}. This can be manipulated to improve the efficiency of the calculation, or simply used to characterize the overall body order and the type of correlations included in the architecture.

\begin{figure}[btp]
    \centering
    \includegraphics[width=0.8\linewidth]{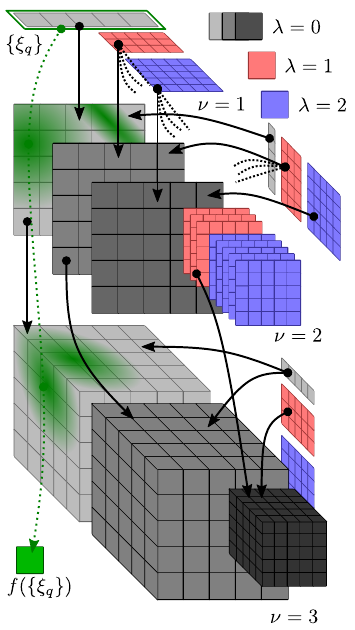}
\caption{\rev{ The figure extends the scheme in Fig.~\ref{fig:cg-iter} to illustrate the construction of a hierarchy of invariant features of order $\nu=1,2,3,\ldots$ by repeated application of the Clebsch-Gordan (CG) iteration~\eqref{eq:gen-cg-iter}. 
Linearly-independent invariants can also arise from intermediate representations with a vectorial, $\lambda>0$ character.
A non-linear function of the $\nu=1$ invariants selects a single direction within the subspace that is generated exclusively by tensor products of the $\lambda=0$ terms (green dotted line), combined with weights that depend on the details of the non-linear function (green shading). \label{fig:scalars}
} }    
\end{figure}

\subsection{Scalar nonlinearitites.}
\label{sub:scalar}
The discussion this far focuses on the case of \emph{equivariant} MP representations that aim to provide a complete basis for symmetric function of the neighbor positions.  Arguments similar to those exposed in Eq.~\eqref{eq:gc-11-model} make it possible to determine the precise form of the function of interatomic distance vectors that is associated with each type of features. 
\rev{
Even though it is convenient to manipulate these representations in their \emph{irreducible} equivariant form, as long as one exclusively uses tensor products to combine them, the higher-order terms can always be brought back into an irreducible form, which, for instance, is exploited by moment tensor potentials\cite{shap16mms}. 
If however one takes an arbitrary non-linear function of the equivariant components, the result cannot be separated into irreducible terms.
The case of scalar (invariant) components is an exception, in that any non-linear function of invariant features is itself invariant. 
Thus, one can enrich the architecture by applying non-linear functions to the scalar component of the equivariants of any type and order. 
It should be noted that doing so eliminates the link of ACDCs with a body-ordered \emph{linear} expansion of the target property.
}
As a simple example, Behler-Parrinello neural networks\cite{behl-parr07prl} compute atom-centered symmetry functions that are equivalent to $\rep|\frho_i^1>$ and $\rep|\frho_i^2>$, and use them as inputs to a feed-forward neural network, that is essentially a non-linear function with adjustable parameters. \rev{ This corresponds to a non-linear functional form that can incorporate some, but not necessarily all, higher-order correlations.\cite{pozd+20prl}} 

To see what these non-linear terms do, we start from a vector of invariants $\bfeat(A_i) = \{\feat_q(A_i)\} = \{\rep<q||A_i>\}$ (e.g. $\{\rep<q||\frho_i^1>\}$, $\{\rep<q||\frho_i^2>\}$, or any other representation such as $\{\rep<q||\frho_i^{[1\leftarrow 1]}>\}$) and apply Eq.~\eqref{eq:gen-cg-iter}, with all terms truncated to $\lambda=0$. This generates all possible products of the initial features $\rep<q_1; q_2||A_i^{\otimes 2}>\propto \rep<q_1||A_i>\rep<q_2||A_i>$.
In terms of the feature vector, this corresponds to $\{\feat_1^2, \feat_1 \feat_2, \feat_2^2, \ldots\}$. Repeating the iteration $k$ times generates all monomials of total order $k$ that combine powers of $\{\feat_q\}$, e.g. $\feat_1^{3}\feat_2^2\feat_9^{k-5}$. 
Comparing with the power series expansion of an analytical function of $\bfeat(A)$
\begin{equation}
f(\bfeat) = \sum_{k_1 k_2 \ldots}  \frac{f^{(k_1,k_2,\ldots)}(0)}{k_1!, k_2!, \ldots} \feat_1^{k_1} \feat_2^{k_2} \ldots,
\end{equation}
one sees that computing $f(\bfeat)$ is equivalent to repeating the $\lambda=0$-truncated iteration to infinity and projecting the resulting features on the expansion coefficients. 
Given that $f$ (and hence the coefficients) usually depends on adjustable parameters, architectures that include scalar functions escape the systematic classification that we propose for equivariant frameworks that are entirely based on the angular momentum iteration. 
Still, the spirit is clear: each non-linear scalar function brings in infinite body-order correlations, that are however restricted to $\lambda=0$ iterations and projected along a single adjustable direction in the infinite-body-order feature space.
\rev{ In other terms, scalar nonlinearities provide a \emph{computationally efficient} way to incorporate high-order products of the scalar components of the different types of ACDCs. However, they are not always more expressive than lower-order equivariant models because (1) each non-linear function picks up a single linearly independent component of the high-order correlation space, %
and (2) they miss entirely all the components of the correlation space that arise from the combinations of $\lambda>0$ equivariants (Figure~\ref{fig:scalars}).  } 

\rev{
\subsection{Relation to equivariant networks}

During the last few years, a class of NNs has been developed that is based on the use of rotationally equivariant expressions as their fundamental building blocks, operating on node and edge descriptors that are computed on a basis of spherical harmonics.  
Models of this type rely heavily on the use of CG iterations analogous to Eq.~\eqref{eq:gen-cg-iter}, and therefore have the most direct relation with the MP-ACDC formalism, up to a one-to-one correspondence when considering frameworks that do not introduce any kind of scalar non-linearities.
The essential ingredient is the construction of contracted message-passing terms, akin to Eq.~\eqref{eq:mp-simple}. In most frameworks, this operation is split into two parts: the construction of a neighbor sum of messages
\begin{equation}
    \!\!\!\!\!\rep|\rho_{{i}}^{\otimes [0\leftarrow \nu_1]}> \!=\!  \sum_{i_1} \! \rep|\rho_{i_1}^{\otimes \nu_1}> \!\otimes \!\rep|\br_{i_1 i}> \label{eq:mp_neighbors_sum},
\end{equation} 
and the interaction of the contracted message with the central atom
\begin{equation}
    \rep|\rho_{{i}}^{\otimes [\nu\leftarrow \nu_1]}> = 
    \rep|\rho_i^{\otimes \nu}> \!\otimes\!\! \rep|\rho_{{i}}^{\otimes [0\leftarrow \nu_1]}>.
    \label{eq:self_cg_interaction}
\end{equation}
}
\rev{
Given that symmmetry considerations greatly restrict the range of permissible operations, various equivariant frameworks are remarkably similar. The most significant differences lie in the strategy used to avoid the exponential increase in the dimensionality of the tensor-product space and make the NN practically computable. 
To establish a direct connection, therefore, it is necessary to explicitly describe the basis used to express the various equivariant terms.
Both $\rep|\rho_i^{\otimes \nu}>$ and $\rep|\rho_{i_1}^{\otimes \nu}>$ can be written in their irreducible form (we neglect the parity index, for simplicity) and enumerated by a feature index $q$, e.g. $\rep<q||\frho[l m]_i^\nu>$. For $\nu=1$, the $q$ index corresponds to radial functions that discretize the spherically-averaged distribution of neighbors. The pair vector term can also be discretized in an irreducible equivariant form as $\rep<q||\overline{\br_{i_1i}; l m} >=\rep<r_{i_1i}||q>\rep<\brhat_{i_1i}||lm>$. 
Thus, Eq.~\eqref{eq:mp_neighbors_sum} can be written as
\begin{multline}
 \rep<q_1 l_1; q_2l_2||\overline{\rho_{{i}}^{\otimes [0\leftarrow \nu_1]}; \lambda \mu}>=\\
\sum_{i_1 m_1 m_2}
\rep<q_1||\overline{\rho_{{i_1}}^{\otimes\nu_1}; l_1m_1}>
\rep<q_2||\overline{\br_{i_1 i}; l_2m_2}>
\cg{l_1m_1}{l_2m_2}{\lambda\mu}\label{eq:mp_base_block}
\end{multline}
following Eq.~\eqref{eq:gen-cg-iter}. 
In practical implementations, one can perform first the sum over the neighbors (as done in tensor field networks\cite{thomas2018tensor}) or perform first the CG product, as we do here. 
The interaction layer (\ref{eq:self_cg_interaction}) adds one more feature index and two more angular momentum indices, e.g.  $\rep<q_1l_1;q_2l_2;q_3l_3 k||\frho[\lambda\mu]_i^{[1\leftarrow 1]}>$.
Iterating Eq.~\eqref{eq:mp_neighbors_sum} and~\eqref{eq:self_cg_interaction} exponentially increases the size of the feature vector.
This is precisely the same issue that is observed in high-order ACDC frameworks,\cite{will+19jcp,drau19prb} and is usually addressed by a contraction/selection of the features, based on either heuristic\cite{vand+20mlst} or data-driven\cite{imba+18jcp,niga+20jcp} considerations. 

In the context of equivariant NNs there are two main approaches to alleviate the exponential increase in the dimensionality of descriptors. The first, \emph{delayed} approach\cite{kondor2018clebsch,niga+20jcp} involves linear contraction steps applied after each CG iteration. In practice, Eq.~\eqref{eq:mp_base_block} is followed by
\begin{equation}
\rep<\tilde{q}||\overline{A; \lambda \mu}>=
\sum_{q_1l_1q_2 l_2} \rep<\tilde{q}; \lambda||q_1l_1;q_2l_2> \rep<q_1l_1; q_2l_2||\overline{A; \lambda \mu}>, 
\label{eq:delayed_contraction}
\end{equation}
where the contraction coefficients $\rep<\tilde{q}; \lambda||q_1l_1; q_2l_2>$ depend on the equivariant channel $\lambda$, and can be determined in an unsupervised manner (e.g. through PCA\cite{niga+20jcp}) or are simply coefficients in the network, optimized by backpropagation. 
The second, \emph{anticipated} approach involves applying a linear transformation to the descriptors \emph{before} performing the CG iteration. 
First, one computes
\begin{equation}
\rep<\tilde{q}; t||\frho[\lambda\mu]_i^\nu> = \sum_q \rep<\tilde q; \lambda; t||q> \rep<q||\frho[\lambda\mu]_i^\nu>\label{eq:self_contraction}
\end{equation}
and
\begin{equation}
    \rep<\tilde{q}; t||\overline{\br_{i_1 i}; \lambda\mu}> = \sum_{q'} \rep<\tilde{q}; \lambda; t||q'> \rep<q'||\overline{\br_{i_1 i}; \lambda\mu}>.\label{eq:pair_contraction}
\end{equation}
Different transformations can be applied at different points in the network, as emphasized by the additional index $t$ in the contraction coefficients $\rep<\tilde q; \lambda; t||q>$ and $\rep<\tilde{q}; \lambda; t||q'>$, and the operation projects the two terms in a space of equal dimensionality even if they are not initially.
Then, Eq.~\eqref{eq:mp_base_block} can be expressed in a more efficient way that does not involve a tensor product
\begin{multline}
 \rep<\tilde{q}; l_1; l_2; t||\overline{\rho_{{i}}^{\otimes [0\leftarrow \nu_1]}; \lambda \mu}>=
\sum_{i_1 m_1 m_2} \cg{l_1m_1}{l_2m_2}{\lambda\mu} \\
\times \rep<\tilde{q}; t||\overline{\rho_{{i_1}}^{\otimes\nu_1}; l_1m_1}>
\rep<\tilde{q}; t||\overline{\br_{i_1 i}; l_2m_2}>.
\label{eq:mp_base_block_contracted}
\end{multline}
Note that \emph{in principle} this form does not entail loss of generality, because one could write the linear transformations~\eqref{eq:self_contraction} and~\eqref{eq:pair_contraction} as a projection in a higher dimensional space, reproducing the full tensor product, which would however defeat the purpose of an anticipated transformation. 

Having expressed the basic building blocks of equivariant neural networks in our notation, we can draw a direct correspondence to different frameworks. 
In the language used in Tensor Field Networks\cite{thomas2018tensor},
Eq.~\eqref{eq:mp_base_block_contracted} describes a pointwise convolution layer, incorporating also a self-interaction layer that corresponds to~\eqref{eq:self_contraction}. The contracted pair term $\rep<\tilde{q}; t||\overline{\br_{i_1 i}; \lambda\mu}>$ corresponds to the rotation-equivariant filter, given that an adaptive linear combination of a fixed radial basis has the same expressive power as a learned radial function -- which is used in NequIP\cite{batzner2021se3}, an atom-based implementation of TFN with an architecture optimized for applications to the construction of interatomic potentials.

All the on-site operations discussed up to now (both the delayed~\eqref{eq:delayed_contraction} and anticipated~\eqref{eq:self_contraction}) correspond to a linear change of basis, and don't increase the expressive power of the descriptor.  
One can however design on-site terms that increase the correlation order relative to a tagged atom, e.g.
\begin{multline}
\rep<q_1 l_1; q_2 l_2||\frho[\lambda \mu]_i^{(\nu_1+\nu_2)}>=
\sum_{m_1 m_2} \cg{l_2 m_1}{l_2 m_2}{\lambda\mu} \\ 
\times \rep<q_1||\frho[l_1m_1]_i^{\nu_1}>\rep<q_2||\frho[l_2m_2]_i^{\nu_2}>.
\end{multline}
This kind of CG iteration -- which is the same as Eq.~\eqref{eq:gen-cg-iter}, that underlies ACDCs -- is used in Clebsch-Gordan Nets\cite{kondor2018clebsch}, where it is combined with delayed contractions to keep the computational cost manageable. 
Other architectures, such as Cormorant\cite{anderson2019cormorant}, combine all three discussed layers. On top of these, SE(3) equivariant Transformers\cite{fuchs2020se} modify the  pointwise convolutional layer by incorporating an attention mechanism,
\begin{multline}
   \rep<q'||\overline{\rho_{{i}}^{\otimes [0\leftarrow \nu]}; \lambda \mu}; \alpha>=\sum_{mm'}\cg{lm}{l'm'}{\lambda\mu} \\
 \times \sum_{i_1} \alpha_{i_1 i}
\rep<q||\overline{\rho_{{i_1}}^{\otimes \nu}; lm}>
\rep<q; l||\overline{\br_{i_1 i} l'm'}>, \label{eq:mp_se3_transformer}
\end{multline}
where the scalar terms $\alpha_{i_1 i}$ indicate the cross-attention coefficients\cite{vaswani2017attention}. These terms are  built by first constructing invariant combinations of the covariant query vector related to the central atom $i$ and key vectors associated  with the neighbor atoms $i_1$, which are there combined with a non-linear soft-max operation.
Since the attention mechanism is not linear, each layer incorporate terms up to infinite body order as discussed in Sec.\ref{sub:scalar}, and so the indices $\nu$ cannot be  rigorously interpreted as a tracker of the body order of the descriptor.
}

\rev{
\subsection{Non-linearities in deep-learning frameworks}\label{sub:deep-l}
} 

\rev{
To discuss the role of non-linearities, and how they can -- at least in principle -- be understood in the language of MP-ACDC let us discuss first the case of neural networks based on invariant, scalar attributes of the nodes and edges of the graph, which constitute the earlier, and better established, message passing and graph-convolution neural networks for molecules. 
A general \emph{invariant} message-passing architecture, as formalized by Gilmer et al.\cite{gilm+17icml}, amounts to steps in which node features $\mathbf{h}_i$ and edge features $\mathbf{e}_{ii_1}$ are combined based on the construction of messages
\begin{equation}
\mathbf{m}^{t+1}_{i} = \sum_{i_1\in A_i} \mathbf{M}_t(\mathbf{h}^t_i, \mathbf{h}^t_{i_1}, \mathbf{e}_{ii_1}) \label{eq:gilmer-message}
\end{equation}
and their aggregation to update the node features
\begin{equation}
\mathbf{h}^{t+1}_{i} = \mathbf{U}_t(\mathbf{h}^{t}_{i}, \mathbf{m}^{t+1}_{i})
\label{eq:gilmer-update}
\end{equation}
The linear limit of this architecture can be described as a special case of the MP-ACDC.
The node features $\mathbf{h}^t_i$ can be taken to be any set of scalar features. At the simplest level, one can take $\nu=1$ invariants, i.e. projections of the interatomic distances on a radial basis, which can represent any function of the distance. 
Thus, the linear message function $\mathbf{M}_t$ corresponds to the 2-centers ACDC, restricted to a tensor product of site and edge invariants
\begin{equation}
\rep|\frho_{{ii_1}}^{11}; \lmax=0> =  
\rep|\frho_{i}^{1}> \otimes  \rep|\frho_{{i_1}}^{1}> \otimes \rep|r_{{ii_1}}>
\end{equation}
appropriately discretized on a radial basis. The contraction over $i_1$ yields the MP-ACDC (again, restricted to invariant features)
\begin{equation}
\rep|\frho_i^{[1\leftarrow 1]}; \lmax=0> = \sum_{i_1}
\rep|\frho_{{ii_1}}^{11}; \lmax=0>, 
\end{equation}
while the node update simply increases the body order of the on-site descriptor
\begin{equation}
\rep|\rho_i^{\otimes \nu}>\otimes \rep|\rho_i^{\otimes [\nu_1\leftarrow \nu_2]}> = \rep|\rho_i^{\otimes [(\nu_1+\nu)\leftarrow \nu_2]}>.
\end{equation}

The general, non-linear case amounts to using the pair features $\feat^{(i i_1)}_j = \rep<j||\frho_{{ii_1}}^{11}; \lmax=0>$  as the inputs for a set of scalar functions. If one wants to keep a linear formalism, arbitrary non-linear message function $M_t$ can be approximated by a linear fit using tensor products that approximate the high dimensional scalar correlations space, as discussed in Section~\ref{sub:scalar}.
The entries of $\mathbf{m}_i^{t+1}$ could be written explicitly as
\begin{multline}
\rep<q||\frho_i^{[1\leftarrow 1]}; \lmax=0;\mathbf{M}_t>
=\sum_{i_1} M_{q,t} (\bfeat^{(i i_1)}) =
\\\sum_{i_1} \sum_{k_1 k_2 k_3\ldots} M_{q,t}^{k_1,k_2\ldots} [\feat_1^{(i i_1)}]^{k_1}[\feat_2^{(i i_1)}]^{k_2}\ldots\label{eq:Mqt-series}
\end{multline}
setting $M_{q,t}^{k_1,k_2\ldots}$ to the appropriate values to match the series expansion of $M_{q,t}$.
Even though Eq.~\eqref{eq:Mqt-series} shows how to formulate the message in terms of MP-ACDC, it is clearly an impractical way of evaluating a prescribed non-linear function of  $\bfeat^{(i i_1)}$.
Similar considerations apply to the update functions, Eq.~\eqref{eq:gilmer-update}.
In the non-linear case,  $U_t$ also introduces higher products of $\rep|\rho_i^{\otimes [\nu\leftarrow \nu']}; \lmax=0>$ features, and therefore is akin to the calculation of representations that combine joint information of multiple neighbors, e.g. Eq.~\eqref{eq:contract-two-centers}. The restriction to $\lmax=0$, however, reduces the expressive power of the combination relative to a full tensor product.

\subsection{Extending the framework}

We have demonstrated the direct connections between the MP-ACDC formalism and many established invariant and equivariant neural networks. There are several directions along which the framework can be extended, with qualitative connections to recently proposed neural network architectures that attempt to achieve a better balance between the simplicity of invariant networks and the accuracy of their equivariant counterparts. 

One obvious way to increase the expressive power of invariant deep learning models is to start from higher-order scalar descriptors -- or to incorporate them at later stages of the iteration. This has long been known for NN potentials based on atom-centered symmetry functions\cite{behl-parr07prl,smit+17cs}, which have included, from the very beginning 3-body symmetry functions.
Angular information can be readily included into an ACDC through the discretized 2-center invariant $\rep|\frho_{{ii_1i_2}}^{0}>$ (see Eq.~\eqref{eq:nu0-n2}), as well as its contracted pair $\rep|\frho_{{ii_1}}^{1}>$ or atom-centered $\rep|\frho_{i}^{2}>$ counterparts -- the latter corresponding to the SOAP powerspectrum. 
Several works \cite{klicpera2020directional,choudhary2021atomistic} have reported substantial improvements in the performance of ML models when incorporating information on interatomic angles into message-passing frameworks.
REANN\cite{zhan+21prl} is a particularly interesting example, that iteratively combines three-body invariant features weighted by coefficients describing the corresponding pair of neighbors -- essentially the scalar equivalent of the MP contraction over two decorated neighbors~\eqref{eq:contract-two-centers} -- overcoming some of the limitations of both distance-based MP frameworks and atom-centered 3-body invariants. 
}

\rev{
Returning to equivariant NNs, non-linearities can be included in a similar spirit as for invariant message-passing schemes, increasing the effective correlation order without the cost of a full CG iteration.
A simple way to introduce a non-linear term without interfering with the linear nature of SO(3) covariance involves combining equivariant features with a scaling computed from with invariant non-linear terms, e.g.
\begin{equation}
\label{eq:non-linear-scalar}
\rep<q||\overline{A; \lambda\mu}; f> = f(\{\rep<q'||\overline{A; 00}>\})\rep<q||\overline{A; \lambda\mu}>,
\end{equation}
an idea that was already discussed in the context of ACDCs\cite{will+19jcp}.
Most of the recent architectures that extend equivariant NNs, incorporate some kind of non-linearities, such as the attention mechanism discussed for SE(3) transformers in Eq.~\eqref{eq:mp_se3_transformer}. 
Other noteworthy, recent attempts to improve on equivariant NNs  include: GemNET\cite{klic+21arxiv} -- that introduces a two-hop message passing scheme which is rotationally equivariant through directional representations of interatomic separations defined on the $S^2$ sphere, rather than on the SO(3) group, resulting in spherical convolution layers that depend on angles between interatomic separation vectors;  PAINN\cite{schutt2021equivariant} -- which propagates directional information in Cartesian space, reducing the complexity associated with the construction of a complete equivariant basis set, and can be mapped to MP-ACDCs with inputs restricted to $\lmax=1$; SEGNN\cite{brandstetter2021geometric} -- which extends the equivariant (steerable) node features to include spherical decompositions of tensors representing physical/geometric quantities to perform contractions of type Eq.~\eqref{eq:gen-cg-iter} with the messages, in addition to incorporating non-linearities of the form Eq.~\eqref{eq:non-linear-scalar} in the message and node-update functions. 
Further extensions of the MP-ACDC formalism may be needed to characterize  systematically these extensions, particularly for in regard to the use of non-linear terms.  
}

\section{Examples}
\vspace{-1ex}
We choose \rev{three} simple examples to illustrate the behavior of the MP representations in the context of linear models  (details are given in the \SM).
\rev{We emphasize that we deliberately chose simple examples and small data sets, as we do not intend to perform a benchmark exercise. In the following examples, we compare the behavior of different types of representations, and gather some insights on their strengths and shortcomings when used in a controlled setting. }

\begin{figure}[btp]
\centering
\includegraphics[width=1.0\linewidth]{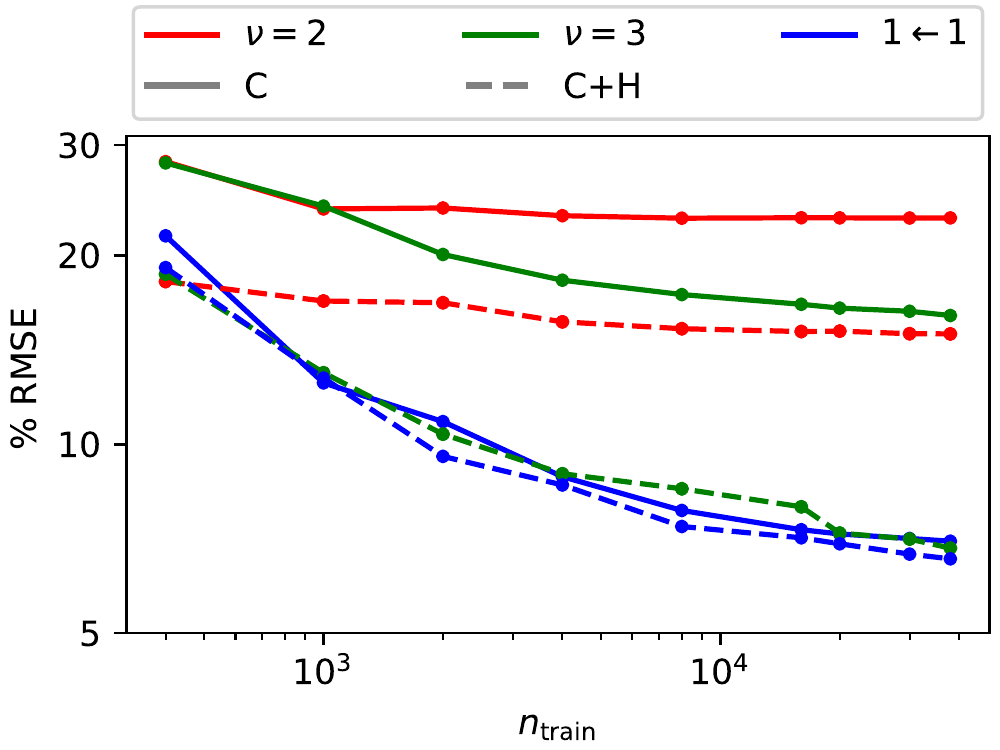}
\caption{Learning curves for the atomization energy of random \ce{CH4} configurations, for different \ce{C}-centered features (full-lines) and for multi-center models using both \ce{C} and \ce{H}-centered features (dashed lines).}
\label{fig:ch4-lc}
\end{figure}

\begin{figure}[btp]
\centering
\includegraphics[width=0.8\linewidth]{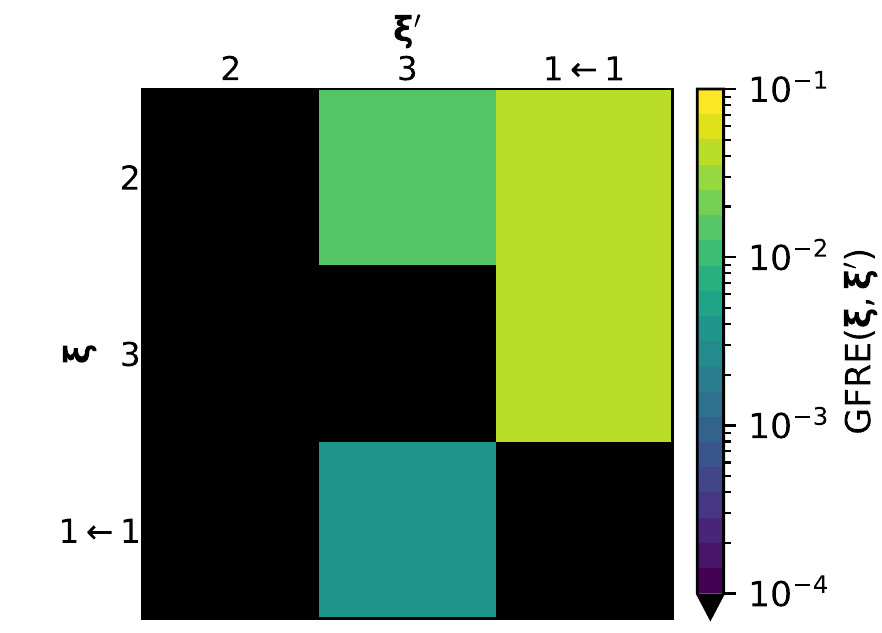}
\caption{\rev{$\text{GFRE}(\bfeat,\bfeat')$ between different types of C-centered features, i.e. the fractional error made when using features $\bfeat(A)$ to linearly predict features $\bfeat'(A)$ for the same configurations.\cite{gosc+21mlst}
}}
\label{fig:methane-c-gfre}
\end{figure}

\subsection{Body-ordered interactions}
\label{sub:methane}
To test the convergence of the body-order expansion we use a dataset of 40,000 random \ce{CH4} configurations\cite{matcloud20a}, with energies computed at the DFT level. We discretize the neighbor density on $\nmax=6$ optimal radial functions\cite{gosc+21jcp}, and $\lmax=4$ angular channels. We set $\rcut=3.5$~\AA{}, so that all atoms are included in each neighborhood.\footnote{The \ce{H} atoms are randomly distributed in a sphere of 3~\AA{} around the central carbon.} To reduce the complexity of the tests and eliminate possible confounding factors we compute the full invariant correlation features up to $\nu=3$, and the $\rep|\frho_i^{{[1\leftarrow 1]}}>$ MP representations, without any data-driven\cite{niga+20jcp} or heuristic contraction.
An interesting observations in Ref.~\citenum{niga+20jcp} is the fact that linear models based on \ce{C}-centered $\nu=4$ features -- that should be complete in terms of expanding a symmetric function of the coordinates of the four \ce{H} atoms -- show saturation of the test error at around 4\%{} RMSE. Models based on $\nu=3$ features saturate at 8\%{} RMSE. 
This saturation is related to the truncation of the basis, and is even more evident here (Fig.~\ref{fig:ch4-lc}) where we use smaller $(\nmax,\lmax)$ to enumerate the full 1-center-3-neighbors correlations, and the $\nu=3$ model cannot improve beyond 16\%{} RMSE (Fig~\ref{fig:ch4-lc}). 

MP representations $[1\leftarrow 1]$ perform dramatically better, reaching a 7\%{} RMSE. This appears to be due to better convergence of the discretization, and not on modulation of the weight of contributions at different distances, which does change the accuracy of the model (see the \SM) but not in terms of the relative performance of $\nu=3$ and MP features. 
Using a multi-center model (Fig~\ref{fig:ch4-lc}, dashed curves) almost entirely eliminates the advantage of MP features over the bispectrum, consistent with the interpretation that the \ce{H}-centered density provides a faster-converging description of the interaction between pairs of hydrogen atoms that are far from the \ce{C}.\cite{niga+20jcp} 
\rev{However, simply having \ce{H}-centered features is not sufficient to obtain a very accurate linear model. Even though $\nu=2$ features also yield more accurate predictions when using a multi-center mode, the improvement is less dramatic than for $\nu=3$ ACDCs, that in the multi-center case are almost equivalent to the MP features.
Even in this very simple test problem the order of the representation, the convergence of the basis and use of a single-centered or multi-centered models are interconnected.  }

\rev{
An alternative test to quantify the relative information content of different representations, independent from the regression target, involves computing the  error one makes when using one type of features $\bfeat$ to  predict a second type $\bfeat'$ for the same structures. 
This construction, recently introduced as the the global feature space reconstruction error\cite{gosc+21mlst} (GFRE) yields a number close to 1 when the target features $\bfeat'$ are linearly independent from $\bfeat$, and close to zero when they can be predicted well.  The GFRE is not symmetric: if $\text{GFRE}(\bfeat, \bfeat') \ll \text{GFRE}(\bfeat', \bfeat)$ one can say that $\bfeat$ is more informative than $\bfeat'$, and vice versa.
Fig.~\ref{fig:methane-c-gfre} demonstrate that all 4-body representations allow predicting, almost perfectly, $\nu=2$ features, and that message-passing ACDCs $[1\leftarrow 1]$ are more informative than their single-center counterpart $\nu=3$, even though the bispectrum does contain a small amount of complementary information, given that $\text{GFRE}([1\leftarrow1], \nu=3)$ is small but non-zero. These results provide further evidence to support the hypothesis that -- for a finite basis -- MP representations may provide a more converged description of interatomic correlations. 
}

\subsection{Electrostatic interactions}
\label{sub:nacl}
\begin{figure}[btp]
    \centering
    \includegraphics[width=1.0\linewidth]{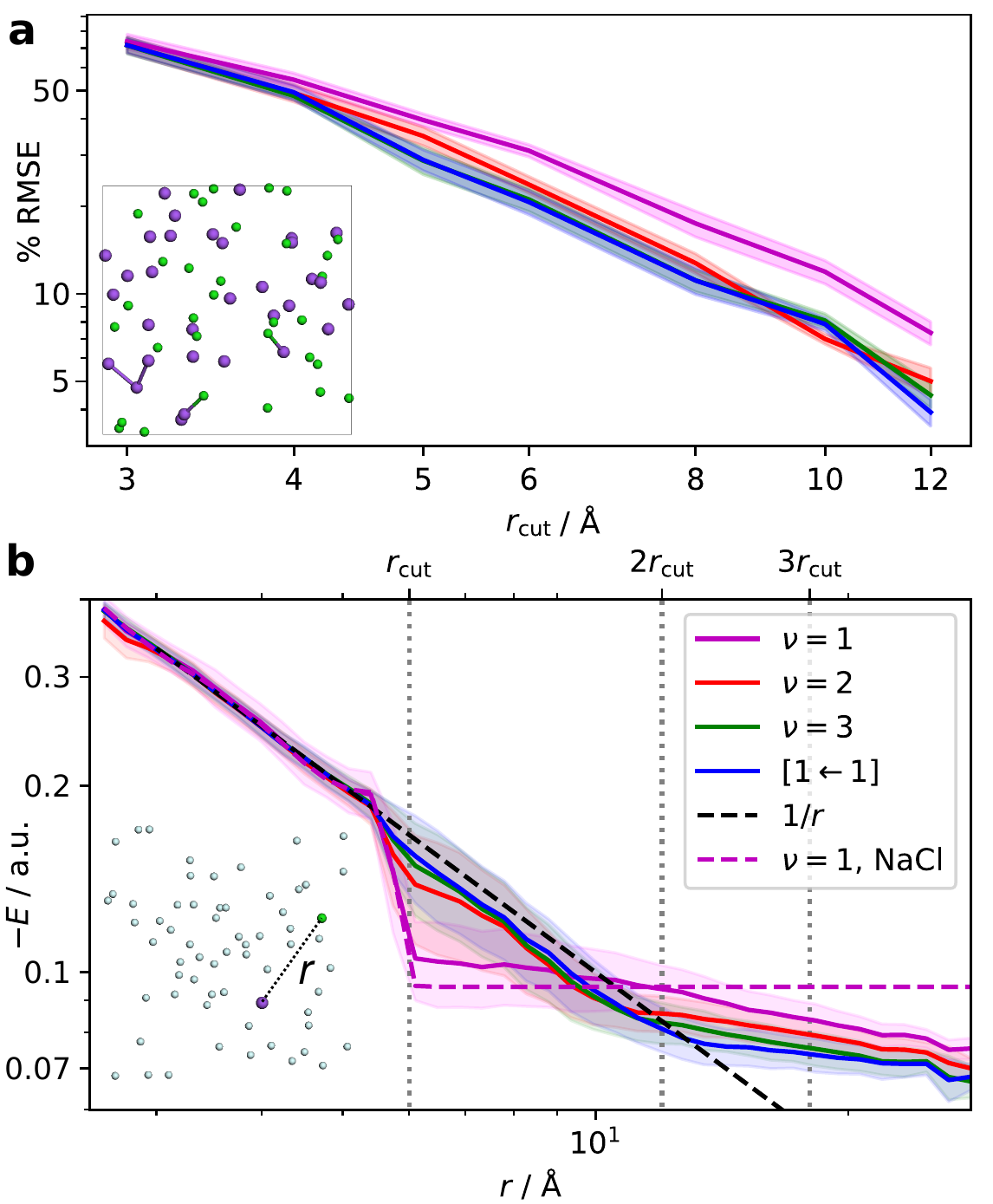}
\caption{(a) Convergence of the $n_{\text{train}}=1800$ validation error for the electrostatic energy of randomized \ce{NaCl} structures, as a function of $\rcut$. The curves correspond to different types of features, following the key in the legend. Shading indicates the standard deviation over three random splits.
(b) Predicted energy of a single \ce{NaCl} pair surrounded by inert atoms as a function of the distance, for linear models built on $\rcut=6$~\AA{} features. The line and the shading indicate the conditionally-averaged mean of predictions for a given distance, and a one-standard-deviation range around it. \rev{The various models yield a test-set RMSE of 0.028 a.u. (NaCl, $\nu=1$), 0.025 a.u. ($\nu=1$), 0.019 ($\nu=2$), 0.017 a.u ($\nu=3$) and 0.016 a.u. ($[1\leftarrow 1]$).}
    }
    \label{fig:nacl-rcut}
\end{figure}

A separate issue is the ability of MP features to incorporate long-range correlations, which we investigate considering a dataset of 2'000 random \ce{NaCl} bulk configurations\cite{gris-ceri19jcp}, with energies computed as the bare Coulomb interactions between $\pm e$ charges. 
This system entails very long-ranged, but perfectly two-body, interactions. Thus, $\rep|\frho_i^1>$ invariant features can learn perfectly well the electrostatic energy, although a very large cutoff is necessary to reduce the error below 10\%{} RMSE (Fig.~\ref{fig:nacl-rcut}a). 
Frameworks that explicitly incorporate a long-range component\cite{gris-ceri19jcp,gris+21cs} provide a more efficient approach to achieve high-accuracy models.

\begin{figure}[btp]
    \centering
    \includegraphics[width=0.85\linewidth]{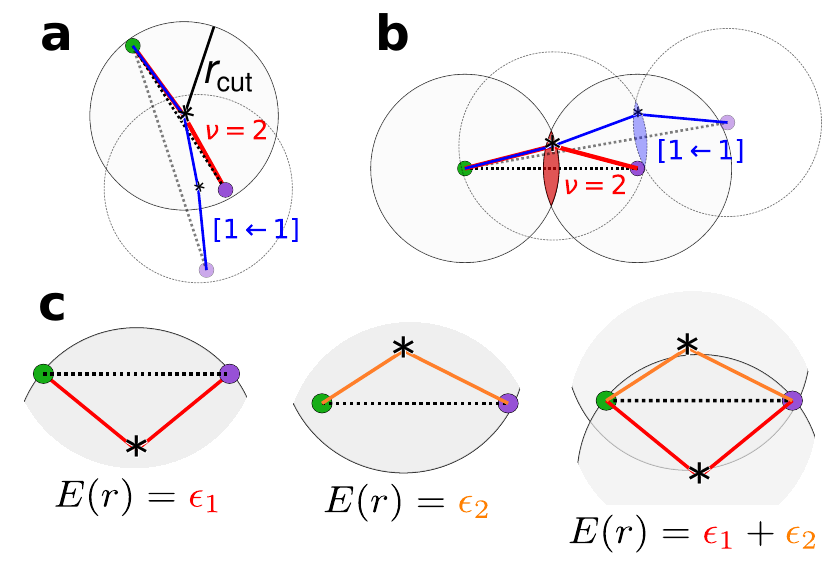}
\caption{(a) Schematic of a configuration for which a $\nu=2$ (or $[1\leftarrow 1]$) representation can encode information on a distance $r>\rcut$;
(b) to have information on the pair, the central (and the decorated neighbor) atoms must lie in very narrow regions, highlighted.
(c) An example of how the same distance can be represented by multiple 3-body terms. Regressing $E(r)$ in terms of the $\chi$-centered contributions, $\epsilon_1$ and $\epsilon_2$, leads to an over-determined system. 
}
\label{fig:lr-schemes}
\end{figure}

It is often stated that  $\nu\ge2$ features contain information on atoms that are up to $2\rcut$ apart\cite{artr+11prb,deri+21cr}. This is formally true, if the two atoms are simultaneously neighbors of a third (Fig.~\ref{fig:lr-schemes}a).
Similarly, $[1\leftarrow 1]$ MP features can in principle describe correlations up to $3\rcut$. Message-passing schemes are often cited as a way to incorporate long-range physics in atomistic ML, and some promising results have been shown for condensed-phase applications\cite{zhan+21prl}. However, Fig.~\ref{fig:nacl-rcut}a clearly shows that -- at least in combination with a linear model -- increasing the range of interactions indirectly is much less effective than just using a large $\rcut$ with two-body features ($\nu=1$).

To better understand this observation, we modify the structures by (1) removing periodic boundary conditions and (2) leaving only one NaCl pair, turning all other atoms to inert \rev{dummy particles, which we label $\chi$} and assume not to contribute to the target energy. Thus, the energy of each structure is just $-E(A)=1/r_{\ce{NaCl}}$.  
A model that uses $\nu=1$ features that describe Na and Cl, \emph{completely ignoring the spectator atoms} yields the expected behavior,  fitting the target almost exactly until $r=\rcut$, and then becoming constant ($\nu=1, \ce{NaCl}$ in Fig.~\ref{fig:nacl-rcut}b).

\rev{One may then investigate what happens when dummy atoms are included in the model. At first, this may seem unnecessary: $\chi$ particles don't contribute to the energy, and in this dataset their positions are almost random. 
However, dummy particle do provide additional information about the Na and Cl atoms that are in their environments -- e.g. reporting on the relative positions of the ions when they are farther than $\rcut$.
}
Simply including dummy atoms within $\nu=1$ features gives the predicted  $E(r)$ curve a small slope for $r>\rcut$: this is because the atom distribution is not entirely random - there is an ``exclusion zone'' of 2.5\AA{} around each atom, and so information on the relative position of \rev{$\chi$} around Na indirectly indicates the possible presence of a Cl atom outside the cutoff.
\rev{Even though overall the validation-set RMSE decreases -- as $\chi$-centered features do indeed incorporate usable information --} the quality of the fit at short $r$ degrades.  
Features with $\nu\ge 2$ (that in theory contain enough information to precisely determine the relative position of Na and Cl up to $2\rcut$) improve the accuracy beyond $\rcut$, but there is a large spread around the target, even if all these results are obtained in the large $n_\text{train}$ limit. 

We believe there are two reasons why the usual argument that $3$-body features extend the range of interactions is too simplistic. 
First, to have some information on $r_{\ce{NaCl}}$ the dummy particles should be placed in a very precise region, which is increasingly unlikely as $r$ approaches $2\rcut$ (Fig.~\ref{fig:lr-schemes}a). Similar arguments apply to $[1\leftarrow 1]$ MP features (Fig.~\ref{fig:lr-schemes}b), that indeed only marginally increase the accuracy of predictions for large $r$.
Second, using a 3-body term to describe the \ce{Na-Cl} interactions may lead to contradictory scenarios (Fig.~\ref{fig:lr-schemes}c). \rev{The  idea of the counterexample is similar to one of the cases discussed in Ref.~\citenum{pozd+20prl}. Consider three structures having the same \ce{Na-Cl} distance, that is larger than $\rcut$, and one or more $\chi$ atoms that are sufficiently close to see the two ions within the cutoff but outside each other's cutoff. 
It is then possible to assign the same energy contribution to different $\ce{Na}-\chi-\ce{Cl}$ geometries. However if a structure has two dummy atoms simultaneously close to the ion pair, with the same relative distances and angles as the previous two structures, one arrives at a contradictory, overdetermined regression. Note that for simplicity we consider only $\chi$-centered descriptors: adding Na and Cl features would introduce $\chi-\ce{Na}-\chi'$ terms, and break the tie, but one can design a more complicated set of structures that would be similarly contradictory.  }
Overall, the analysis of these two toy models suggests that a fully equivariant MP construction is not necessarily an efficient route to describe long-range/low-body-order interactions, even though it provides, in principle, a systematically convergent basis to express them.

\subsection{\rev{Equivariant predictions of molecular dipoles}}

\rev{
As a final example, we demonstrate the extrapolative prediction of molecular dipole moments of small organic molecules. The dataset is taken from Ref.~\citenum{veit+20jcp}, where a similar exercise was performed to compare the performance of a model based on atom-centered equivariant contributions with one based on the prediction of atomic point charges. 
The purpose of this example is two-fold. First, it serves as a demonstration of the symmetry-adapted prediction of a covariant property. Second, given that we train the model on small molecules taken from the QM7 data set\cite{mont+13njp}, and we make prediction on slightly larger structures from the QM9 set\cite{rama+14sd}, the example gives an indication of differences in transferability between the various representations. 
\footnote{We consider exclusively the 6874 QM7 structures that contain only C, H, N, O atoms, and make predictions on 1000 randomly selected QM9 structures, similarly limited to 4 elemental components.} }

\begin{figure}[btp]
\centering
\includegraphics[width=1.0\linewidth]{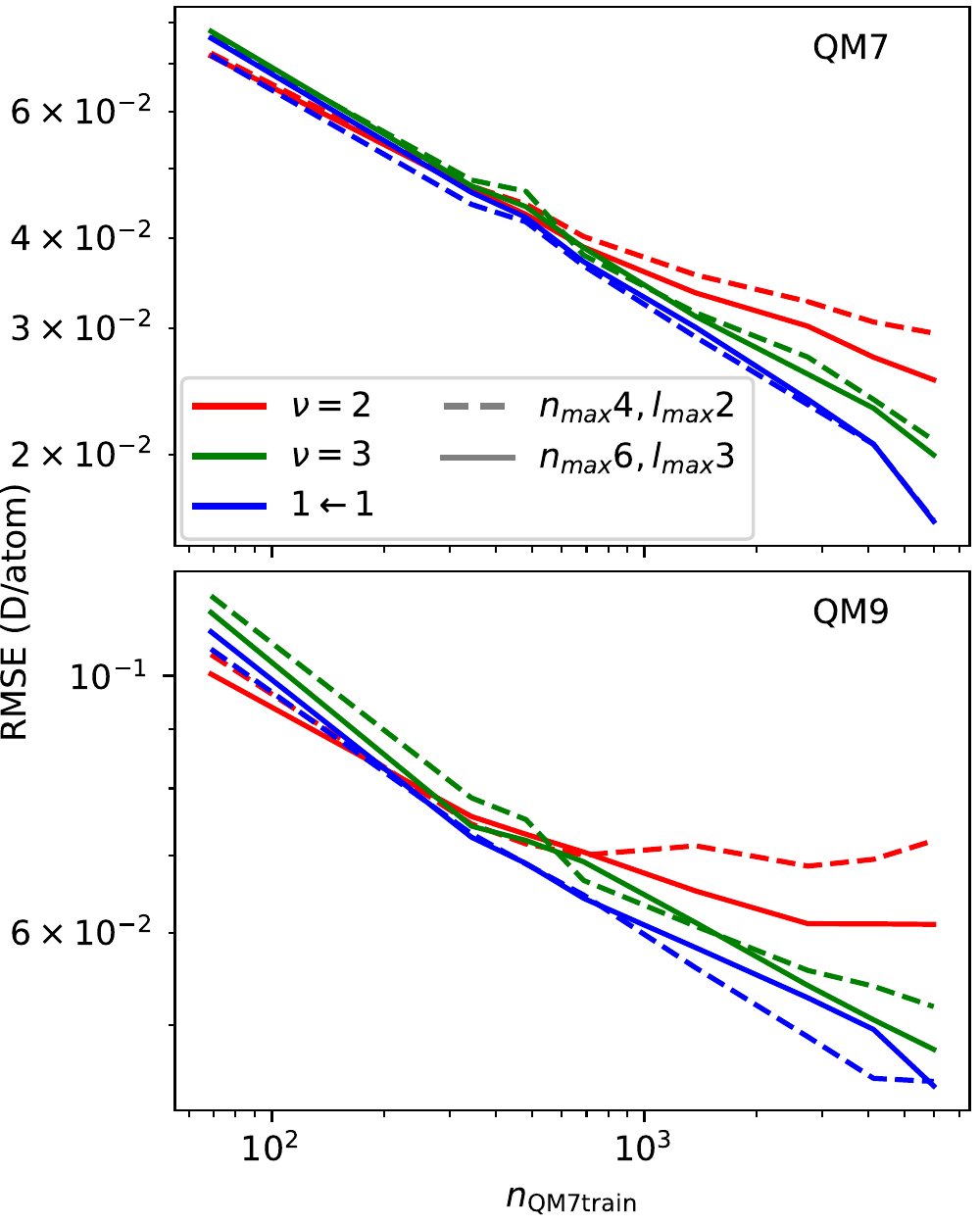}
\caption{\rev{Learning curves for the prediction of molecular dipole models using a linear equivariant model trained on QM7 structures (restricted to CHNO composition), using different types of representations. (top) Validation error on 800 hold-out QM7 structures. (bottom) Error for extrapolative prediction on 1000 larger molecules taken from QM9. }
}
\label{fig:qm7-qm9_lc}
\end{figure}

\rev{
Figure~\ref{fig:qm7-qm9_lc} shows the learning curves of linear models built on $\nu=2$, $\nu=3$ and $[1\leftarrow1]$ features, both for a validation subset of QM7, and 1000 of larger QM9 molecules. We also compare two different levels of discretization of the density expansion (as for \ce{CH4}, we don't truncate the expansion, to keep the setup as simple as possible). 
$\nu=2$ features show inferior performance to those observed in previous work, which is explained both by the use of a larger basis and (more importantly) a non-linear kernel, while the 4-body representation performs much better both for QM7 and for extrapolation to QM9, with performances comparable to the non-linear kernel model in Ref~\citenum{veit+20jcp}. 
The $[1\leftarrow 1]$ MP-ACDC features are remarkably well-behaved. Even with the very coarse $(\nmax=4, \lmax=2)$ discretization, learning curves on QM7 do not saturate, and also the QM9 extrapolation error decreases steadily with $n_\text{train}$. 
Despite having restricted our tests to a linear model, the accuracy reached on QM9 is better that was achieved in Ref.~\cite{veit+20jcp} with non-linear kernels and a hybrid model using both atomic dipoles and partial charges -- even though it must be stressed that the models in Ref.~\cite{veit+20jcp} included also sulfur-containing molecules.    

}

\vspace{-3ex}
\section{Conclusions}
\vspace{-1ex}
The formalism we introduce in this work extends the atom-centered density-correlation representations, and their link with linear models of functions of interatomic displacements, to the case in which information on multiple environments is combined in a similar symmetry-adapted construction. 
The scheme captures the key ingredients of equivariant message-passing ML architectures, which however usually include additional scalar terms that make them more flexible but prevent establishing a clear correspondence with body-ordered expansions.
We illustrate the behavior of the simpler MP representations on three elementary examples -- the first focusing on the description of short-range body-ordered interactions, the second on long-range properties, and the third to covariant targets and extrapolative power. 
The examples suggest that MP representations can incorporate information on the position of atoms more efficiently than single-atom-centered features, even though this advantage is reduced when using a multi-center decomposition of a global target property. 
The analysis of a purely electrostatic learning problem provides insights on the ability of MP features to incorporate long-range correlations. The formal observation that correlations between multiple neighbors extend the range of interactions beyond the cutoff appears to be of limited practical relevance. 
\rev{MP-ACDCs also perform very well in a more realistic application to small organic molecules, and confirms their better behavior in the limit of a heavily truncated discretization, compared to single-center features of the same body order.  }

\rev{On a formal level, we show how invariant and equivariant deep learning frameworks can be interpreted in terms of time-tested atom-centered descriptors.}
We expect that this formalism will help rationalize and classify geometric message-passing machine-learning schemes, that are often touted as ``features-free'' approaches, but usually entail the \emph{ad hoc} construction of heavily-engineered architectures.  From this point of view, the main challenge is to study systematically the role played by scalar terms -- that correspond to infinite correlation order, but restricted to powers of the input invariant features -- and the iteration of the MP step, that also requires introducing contractions to limit the size of the feature vectors. 
A better understanding of the link between different kinds of ML architectures and systematically-converging representations, and between representations and descriptions of physically-motivated components of the targets, will help in the construction of more accurate, efficient and transferable machine-learning models of atomic-scale problems, as well as all applications that rely on a geometric description of point clouds. 

\section*{Data availability}
\rev{
The software used to compute the new representations in this work is available online on Zenodo\cite{Zenodo}. All data and scripts used for the examples in section II are available online on the Materials Cloud archive\cite{MaterialsCloudArchive}.
}

\begin{acknowledgments}
JN and MC acknowledge support by the NCCR MARVEL, funded by the Swiss National Science Foundation (SNSF) and the European Research Council (ERC) under the European Union’s Horizon 2020 research and innovation programme (grant agreement No 101001890-FIAMMA). GF and SP acknowledge support by the Swiss Platform for Advanced Scientific Computing (PASC). Discussions with Alexander Goscinski and Andrea Grisafi are gratefully acknowledged.

\end{acknowledgments}

\onecolumngrid
\clearpage
\newpage

\end{document}